\pdfoutput=1
\documentclass[11pt]{article}

\usepackage[final]{acl}

\usepackage{times}
\usepackage{latexsym}

\usepackage[T1]{fontenc}
\usepackage[utf8]{inputenc}

\usepackage{microtype}

\usepackage{inconsolata}

\usepackage{rotating}
\usepackage{float}
\usepackage{times}
\usepackage{latexsym}
\usepackage{tablefootnote}
\usepackage{graphicx}
\usepackage{booktabs}
\usepackage[utf8]{inputenc}
\newcommand{\ignore}[1]{}
\newcommand{\squishlist}{
 \begin{list}{$\bullet$}
  { \setlength{\itemsep}{0pt}
     \setlength{\parsep}{2pt}
     \setlength{\topsep}{2pt}
     \setlength{\partopsep}{0pt}
     \setlength{\leftmargin}{1em}
     \setlength{\labelwidth}{1em}
     \setlength{\labelsep}{0.4em} } }

\newcommand{\squishend}{
  \end{list}  }

\usepackage[compact]{titlesec}
\usepackage{url}
\usepackage{cleveref}

\usepackage{tabularx}
\usepackage{geometry}
\usepackage{setspace}
\usepackage{colortbl}
\usepackage{xcolor}
\definecolor{darkgreen}{RGB}{0,100,0}
\usepackage{siunitx}  %for aligning numbers in table

\usepackage{array,multirow}

\title{Beyond Flesch-Kincaid: Prompt-based Metrics Improve \\Difficulty Classification of Educational Texts}

\author{
    Donya Rooein$^{1}$, 
    Paul Röttger$^{1}$, 
    Anastassia Shaitarova$^{2}$, 
    \textbf{Dirk Hovy}$^{1}$ \\ \\
    {$^1$Bocconi University, $^2$University of Zurich} \\
    \texttt{\{donya.rooein, paul.rottger, dirk.hovy\}@unibocconi.it,}\\
     \texttt{anastassia.shaitarova@uzh.ch}
}

\begin{document}
\maketitle
%  New Abstract
\begin{abstract}
Using large language models (LLMs) for educational applications like dialogue-based teaching is a hot topic.
Effective teaching, however, requires teachers to adapt the difficulty of content and explanations to the education level of their students.
Even the best LLMs today struggles to do this well.
If we want to improve LLMs on this adaptation task, we need to be able to measure adaptation success reliably.
However, current \textsc{Static} metrics for text difficulty, like the Flesch-Kincaid Reading Ease score, are known to be crude and brittle.
We, therefore, introduce and evaluate a new set of \textsc{Prompt-based} metrics for text difficulty.
Based on a user study, we create \textsc{Prompt-based} metrics as inputs for LLMs. They leverage LLM's general language understanding capabilities to capture more abstract and complex features than \textsc{Static} metrics.
Regression experiments show that adding our \textsc{Prompt-based} metrics significantly improves text difficulty classification over \textsc{Static} metrics alone.
Our results demonstrate the promise of using LLMs to evaluate text adaptations to different education levels.
\end{abstract}

%%%%%%%%%%%%%%%%%%%%%%%%%%%%%%%%%%%%%%%%%%%%%%%%%%%%%%%%%%%%%%%%%%%%%%%%%%%%%%%%%%%%%%
%%%%%%%%%%%%%%%%%%%%%%%%%%%%%%%%%%%%%%%%%%%%%%%%%%%%%%%%%%%%%%%%%%%%%%%%%%%%%%%%%%%%%%
\section{Introduction}

Large language models (LLMs) today can answer wide-ranging questions and explain complex concepts with high accuracy \citep{chung2022scaling,openai2023gpt4}. 
This development has motivated explorations into their uses for education, ranging from automated student assessment and personalised content to dialogue-based teaching \cite{upadhyay2023improving,healthcare11060887,yan2023practical,hosseini2023exploratory}.

% Controllable text generation is a task in which LLMs could potentially excel~\citep{pu2023chatgpt}.

\begin{figure}[ht!]
    \centering
    \includegraphics[width=\columnwidth]{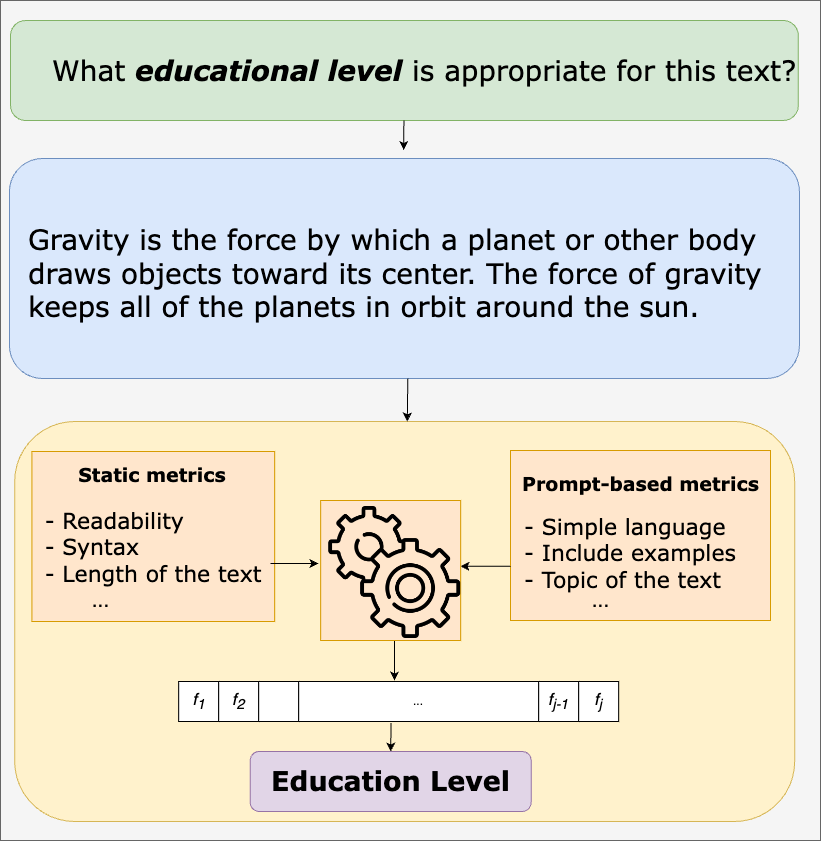}
    \caption{Schematic overview of our approach to text difficulty classification.
    We calculate relevant \textsc{Static} and \textsc{Prompt-based} metrics for a given input text.
    Either or both metrics are then fed into a regression classifier that makes a final classification.}
    \label{fig:QA-model-approach}
\end{figure}

Effective teaching requires that the difficulty of content and explanations is tailored to the education level of the students.
Human teachers are trained to do this, and adjust their material and style without much prompting.
However, this adaptation is not just the adjustment of one variable.
It is a complex undertaking, touching upon lexicon, syntax, pragmatics, and semantics.
% \footnote{\url{https://spaceplace.nasa.gov/what-is-gravity}}
Improving the ability of LLMs to adapt their outputs to different levels of education is therefore crucial to unlocking their usefulness for education.
One of the most basic requirements to achieve this goal is a way to measure adaptation success.

Measuring whether a given output is appropriate for a given level of education, however, is a very difficult task.
Existing \textsc{Static} metrics, like the Flesch-Kincaid Reading Ease score~\citep{flesch1948new}, are based on simple formulas, heuristics, and word counts.
They share the brittleness of all heuristic approaches and are known to be noisy measures of text difficulty at best.
Also, these metrics were developed for longer-form explanations, like those found in textbooks, rather than dialogue-style teaching. Due to their reliance on counts, their estimates are unreliable in shorter formats.
We need better metrics to make improvements on the adaptability of LLMs to education levels measurable.
Only when we can measure improvements can we make tangible progress in leveraging LLMs for educational applications.\footnote{Similarly, metrics like BLEU \cite{papineni-etal-2002-bleu}, ROUGE \cite{lin-2004-rouge}, and BLANC \cite{recasens2011blanc}, among others, kickstarted and sustained the development of automated approaches to machine translation, summarization, and coreference resolution, respectively.}

As an alternative to \textsc{Static} metrics, we can use classifiers to predict the educational level of a given text. They generalize better and can be applied to texts of varying lengths. However, these classifiers are expensive to train and require more training data than we usually have for a niche domain like educational purposes. 
Similarly, human assessment of difficulty may provide a gold standard, but it is expensive to collect and, like all annotation tasks, suffers from disagreement.

In this paper, we introduce and evaluate a new set of \textsc{Prompt-based} metrics for text difficulty as complements to existing \textsc{Static} metrics.
\textsc{Prompt-based} metrics are LLM prompts that exploit the general language understanding capabilities of LLMs to capture more abstract features of educational texts than \textsc{Static} metrics.
For example, LLMs can flexibly classify the topic of a text, which is one adaptation technique used by teachers to adjust the content which called curriculum compacting in pedagogy \cite{stamps2004effectiveness}. This would be difficult to do with \textsc{Static} approaches.

% whether a given text uses a metaphor or not (which is one of the adaptation techniques used by teachers to adjust content to higher education levels).
% This would be difficult to do with \textsc{Static} approaches.

We develop our selection of \textsc{Prompt-based} metrics based on a user study, where we ask a group of university students to 1) assess the difficulty of educational texts and explain their reasoning, and 2) come up with prompts for an LLM to change the difficulty of a given text.
We then translate the qualitative findings from both parts of the study into concrete LLM prompts that serve as \textsc{Prompt-based} metrics. % -- like the metaphor example above.
We incorporate prompts from other studies to manage text readability with LLMs~\citep{imperial2023flesch,gobara2024llms}.
We evaluate the ability of our new \textsc{Prompt-based} metrics to measure text appropriateness for different education levels with a series of regression experiments. 

While \textsc{Prompt-based} metrics perform on par or better than zero-shot and few-shot LLM classifiers, they are less useful for text difficulty classification by themselves than \textsc{Static} metrics.
However, combining \textsc{Prompt-based} and \textsc{Static} metrics significantly improves performance.
This suggests that \textsc{Prompt-based} metrics capture relevant signals beyond those captured by the large number of \textsc{Static} metrics.
% Therefore, the use of \textsc{Prompt-based} metrics in addition to \textsc{Static} metrics appears to be a promising direction for improving text difficulty classification.

%  Added 
A combination of \textsc{Static} and \textsc{Prompt-based} metrics also provides a deeper understanding of the key metrics or features that influence complexity than classifiers could. Additionally, the factors that contribute to complexity in a scientific text differ from those in a medical or a legal document. By considering a range of metrics, we can develop more accurate domain-specific measures. 
Our multifaceted approach allows us to break down complexity into its basic components, such as its appropriateness for different education levels, lexical or syntactic complexity, thematic topics, and text readability. 

Overall, \textsc{Prompt-based} metrics empower educators to develop more effective content development strategies with LLMs to engage learners of all levels and backgrounds. We could have directly trained classifiers; however, this approach would not have enabled us to identify the most relevant metrics.

% and let them comprehend the material effectively.

\paragraph{Contributions}
\begin{enumerate}
    \item We conduct a user study to motivate the creation of novel \textsc{Prompt-based} metrics of text difficulty for educational texts (\S\ref{sec:user_study}).
    \item We show in a series of regression experiments that these \textsc{Prompt-based} metrics hold additional value for text difficulty classification beyond what \textsc{Static} metrics can capture (\S\ref{sec:reg}).
    \item By leveraging the interpretability of our regressions, we highlight the relative importance of individual \textsc{Static} and \textsc{Prompt-based} metrics (\S\ref{sec:results}).
\end{enumerate}

\section{User Study}
\label{sec:user_study}
Our \textsc{Prompt-based} metrics for text difficulty are prompts based on the results of a one-day user study we ran with a group of university students in November 2023.

%Evaluating the educational level of the scientific contents is a multifaceted task that requires metrics to measure their suitability for different learners comprehensively. Previous research studied diverse metrics, including readability scores, which assess sentence and word complexity to determine text's accessibility to readers at various proficiency levels~\cite{solnyshkina2017evaluating}.
%Furthermore, user feedback and comprehension assessments, such as user surveys, provide invaluable insights into how effectively learners engage with and understand the material. 
%We use surveys and interactive assessments to collect critical features in controlling the complexity of educational content. By actively involving students in the research process, the resulting metrics are more likely to align with their needs, thus enhancing the accuracy and relevance of educational content assessments.

%%%%%%%%%%%%%%%%%%%%%%%%%%%%%%%%%%%%%%%%%%%%%%%%%%%%%%%%%
\subsection{Study Design}

The user study consisted of two main parts.

In the first part of our study, we asked participants to review 60 educational texts randomly sampled from the ScienceQA dataset~\cite {lu2022learn}.
Each text consists of a question (e.g., ``What is the mass of a dinner fork?'') with answer choices (``70 grams or 70 kilograms'') and a longer-form explanation of the solution.
All texts we select here are authentic educational materials from the social, natural, or language sciences in schools.
Participants were tasked with a) labeling the education level of each text as appropriate for either elementary school, middle school, or high school and b) explaining the reasoning behind their choice in a short, free-text answer.

In the second part of our study, we asked participants to rewrite scientific text explanations, also sampled from ScienceQA, to be appropriate for different education levels, with the help of an LLM -- in this case, ChatGPT.
For example, participants were asked to rewrite a middle school explanation of thermal energy at the elementary and high school levels with the help of prompts.
We recorded the prompts they used to get ChatGPT to accomplish the adaptation for them.
Thus, we collected prompts that are used both for text \textit{simplification} and for text \textit{complexification}.

%%%%%%%%%%%%%%%%%%%%%%%%%%%%%%%%%%%%%%%%%%%%%%%%%%%%%%%%%
\subsection{Study Participants}
We ran our study as part of a hackathon at the University of Zurich.
There were seven participants aged between 21 and 31 years.
Four participants were female, three male. 
All participants were students at Department of Computational Linguistics from University of Zurich, enrolled at the time in programs specializing in computational linguistics, computer science, and AI.
Five were studying for a bachelor's degree and two for a master's degree.
The participants held prior educational degrees from school systems across five different countries.
Their native languages include English, Italian, German, Greek, and Ukrainian.
They self-reported their English language proficiency at C1 and C2 levels. 
Participants were compensated in study credits that could be counted towards completing their program.

%%%%%%%%%%%%%%%%%%%%%%%%%%%%%%%%%%%%%%%%%%%%%%%%%%%%%%%%%
\subsection{Study Results}

The first task of our study yielded 276 classification labels together with their corresponding descriptive justifications. These include 120 label-explanation pairs for middle school texts, 89 for high school, and 67 for elementary school texts. In the second task of our study, we collected 103 prompts for text simplification and complexification.
We share illustrative examples of classifications, explanations, and prompts in Appendix \ref{appendix-users}.

In the next section, we use the qualitative results from our study to motivate the construction of novel \textsc{Prompt-based} metrics for text appropriateness for various education levels.

%%%%%%%%%%%%%%%%%%%%%%%%%%%%%%%%%%%%%%%%%%%%%%%%%%%%%%%%%%%%%%%%%%%%%%%%%%%%%%%%%%%%%%
%%%%%%%%%%%%%%%%%%%%%%%%%%%%%%%%%%%%%%%%%%%%%%%%%%%%%%%%%%%%%%%%%%%%%%%%%%%%%%%%%%%%%%
\section{Metrics for Text Difficulty}

%%%%%%%%%%%%%%%%%%%%%%%%%%%%%%%%%%%%%%%%%%%%%%%%%%%%%%%%%
\subsection{Prompt-based Metrics}
\label{sec:Prompt-based}

Since the metrics we introduce are based on the prompts of language models rather than discrete heuristics, we refer to them as `\textsc{Prompt-based}' to distinguish them.
The goal of the \textsc{Prompt-based} metrics we develop is to capture more abstract features of educational texts than would be possible with \textsc{Static} metrics, which typically focus on individual words and their statistics.

\begin{figure}[h]
    \centering
    \includegraphics[width=\columnwidth]{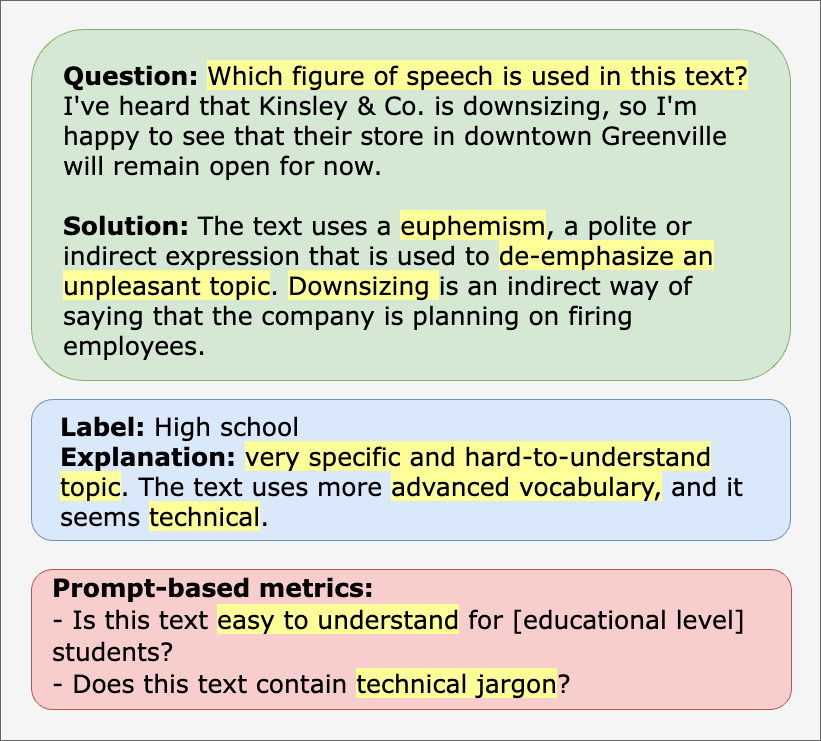}
    \caption{An illustrative example of the \textsc{Prompt-based} metric process. The green box contains the education text from the ScienceQA dataset. The blue box shows the predicted educational level and the explanation. The red box contains the \textsc{Prompt-based} metrics based on the sample.}
    \captionsetup{justification=centering}
    \label{fig:example}
\end{figure}

\begin{figure}
    \centering
    \includegraphics[width=\columnwidth]{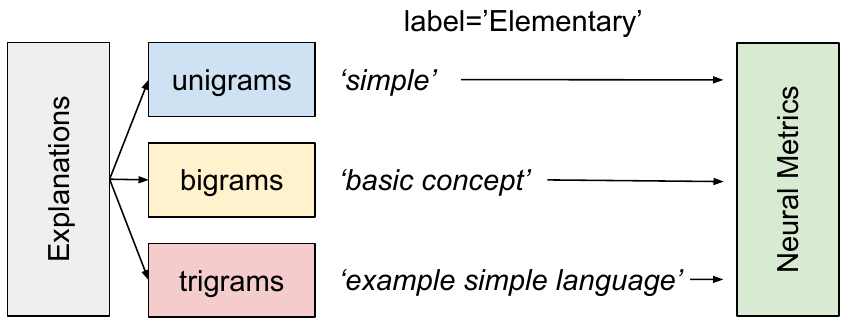}
    \caption{High-level view of the derivation process for the \textsc{Prompt-based} metrics using n-gram frequencies. Function words are excluded.}
    \label{fig:neuralMetrics_highview}
\end{figure}

We derive our \textsc{Prompt-based} metrics from the results of our user study. 
Figure~\ref{fig:example} shows an illustrative example of our derivation process.
We consider users' explanations for why they consider a specific educational text to be of elementary, middle, or high school level difficulty.
Then, we identify recurring attributes and other explanation features that several users mention to reflect them in \textsc{Prompt-based} metrics. We examine the distributions of unigrams, bigrams, and trigrams across all three labels, excluding function words (see Figure \ref{fig:neuralMetrics_highview}). Some of the most frequent unigrams for the elementary level include \textit{simple, basic, elementary}; for the high school level, \textit{high, complex, concepts}; and for 
the middle school level, \textit{explicit, explanation, middle}. 

We qualitatively assessed the n-gram distributions, considering both frequencies and topic appropriateness, before finalizing the query construction. Each \textsc{Prompt-based} metric is a simple yes-no question, which we use to prompt the LLMs. These metrics encompass the most frequent unigrams and less common bigrams and trigrams derived from the findings of our study.

While, \newcite{gobara2024llms} demonstrate a correlation between readability scores of LLM-generated texts in education and human assessments, \newcite{imperial2023flesch} indicate challenges in LLMs effectively adjusting the readability of text. We construct 63 \textsc{Prompt-based} metrics using this process.
Each \textsc{Prompt-based} metric relates to either education level (30 metrics), lexical or syntactic complexity (8 metrics), and the topic of the text at hand (10 metrics). In addition, we include metrics about the text's readability score (15 metrics) based on the work by~\newcite{imperial2023flesch}. The complete list of all our \textsc{Prompt-based} metrics is in Appendix~\ref{appendix-prompts}.

% \begin{table}[h]
% \centering
% \begin{tabularx}{0.47\textwidth}{m{0.27\textwidth}m{0.2\textwidth}}
%     \toprule
%     \textbf{Neural Metric} & \textbf{Category}\\ 
%     \midrule
%     Is this text suitable for an \textbf{elementary school} student? & Education level \\ 
%     \midrule
%     Does this text contain \textbf{technical jargon}? & Lexical/syntactic complexity \\
%     \midrule
%     Is this text about \textbf{math}? & Topic \\
%     \bottomrule
% \end{tabularx}
% \caption{Three example \textsc{Prompt-based} metrics across the three categories of \textsc{Prompt-based} metrics.
% Each \textsc{Prompt-based} metric is a prompt for an LLM.
% }
% \label{tab:neural-metrics}
% \end{table}

%%%%%%%%%%%%%%%%%%%%%%%%%%%%%%%%%%%%%%%%%%%%%%%%%%%%%%%%%
\subsection{Existing Static Metrics}

\textsc{Static} metrics are the baseline we want to improve on.
All \textsc{Static} metrics are based on simple formulas, heuristics, or counts of words and other textual features.
These properties make them simple to apply but limit the conceptual complexity of what they can reasonably measure.
In total, we include 46 \textsc{Static} metrics, selected from those compiled in prior work~\citep{flekova2016exploring, yaneva2019predicting,xue2020predicting,baldwin2021using}.

These metrics encompass a variety of linguistic characteristics, spanning from basic text-level measures like vocabulary size and word frequency to sentence-level attributes such as sentence length and syntactic complexity. Additionally, they take into account the question-answering structure within the input text. In the ScienceQA dataset, each question is paired with its respective solution and corresponding lecture. This segmentation of information across educational levels facilitates the computation of \textsc{Static} features for each section of the question-answer solution and lecture independently.
For the complete list of 46 \textsc{Static} metrics, see Appendix~\ref{appendix-prompts}.

%%%%%%%%%%%%%%%%%%%%%%%%%%%%%%%%%%%%%%%%%%%%%%%%%%%%%%%%%%%%%%%%%%%%%%%%%%%%%%%%%%%%%%
%%%%%%%%%%%%%%%%%%%%%%%%%%%%%%%%%%%%%%%%%%%%%%%%%%%%%%%%%%%%%%%%%%%%%%%%%%%%%%%%%%%%%%
\section{Experiments}

We conduct a series of classification experiments to evaluate the usefulness of our novel \textsc{Prompt-based} metrics for measuring text difficulty. We use a subset of the ScienceQA dataset, which contains question-answer pairs across several topics and education levels.
Specifically, we run multinomial logistic regressions based on \textsc{Static} metrics, \textsc{Prompt-based} metrics, and the combination of the two to evaluate the marginal benefits of our new \textsc{Prompt-based} metrics. 
We also compare these regression approaches to using an LLM for zero-shot and few-shot classification.

%%%%%%%%%%%%%%%%%%%%%%%%%%%%%%%%%%%%%%%%%%%%%%%%%%%%%%%%%
\subsection{Dataset}

All our experiments are based on the ScienceQA dataset~\cite{lu2022learn}.
There are 21,208 texts in ScienceQA.
Each text consists of a question with answer choices, and a longer-form explanation of the solution.
Texts in ScienceQA are classified according to their grade level using the K12 system from the US education system.
We simplify this classification by collapsing the 12-grade levels into just three: elementary school (grades 1 to 5), middle school (grades 6 to 8), and high school (grades 9 to 12).\footnote{\url{https://usahello.org/education/children/grade-levels/}}
From the 21,208 texts in ScienceQA, we sample only those that do not use images in questions or explanations.
We then deduplicate and sample 1,516 texts for each education level to create a balanced dataset of 4,548 texts.
Of these 4,548 texts, we use 3,638 (80\%) for training and 910 (20\%) for evaluation.
To our knowledge, ours is the first use of the ScienceQA dataset for training and evaluating text difficulty classifiers.

%%%%%%%%%%%%%%%%%%%%%%%%%%%%%%%%%%%%%%%%%%%%%%%%%%%%%%%%%
\subsection{LLMs for Prompt-based Metrics}

We use LLMs to compute the 63 \textsc{Prompt-based} metrics described in Section~\ref{sec:Prompt-based}. In principle, any LLM can serve this purpose. With 63 metrics for 4,548 texts, we get 286,524 prompts from each LLM. This amount is prohibitively expensive for paid services like GPT4. Hence, we concentrate on state-of-the-art open LLMs, which we can execute at a low cost: Llama2 \citep{touvron2023llama}, Mistral \citep{jiang2023mistral}, and Gemma \citep{Google2024Gemma7b}. Llama2, launched in July 2023, comprises both pre-trained and fine-tuned LLMs, ranging in size from 7 billion to 70 billion parameters. It has been reported to outperform other open-access LLMs and exhibits capabilities comparable to ChatGPT across various tasks. In this paper, we use Llama2-7b and Llama2-13b. The next model is Mistral-7B, released in September 2023, another open LLM surpassing similar-sized open LLMs. We use Mistral-7b-Instruct-v0.2, which was published in December 2023.

The last model we use is Gemma7b-it, based on the Gemma base model and trained on open-source mathematics datasets. 

% We use LLMs to compute the 63 Neural metrics described in Section~\ref{sec:neural}. 
% In principle, any LLM can be used for this purpose.
% With 63 metrics for 4,548 texts, we need to collect responses to 286,524 prompts from each LLM.
% Therefore, we focus on state-of-the-art open LLMs, which we can run at low cost:
% Llama2 \citep{touvron2023llama}, Mistral \citep{jiang2023mistral}, and Gemma\citep{Google2024Gemma7b}.
% Llama2, launched in July 2023, is a collection of both pre-trained and fine-tuned LLMs, with sizes ranging from 7 billion to 70 billion parameters. It has been reported to outperform other open-access LLMs and demonstrates capabilities comparable to ChatGPT across various tasks. In this paper we use Llama2-7b and llama2-13b. The next model is  Mistral-7B, which released in September 2023, and is another open LLMs that surpasses similar-sized open LLMs. We use Mistral-7b-Instruct-v0.2, which was published in December 2023. The last model we use is Gemmma7b-it which is trained on mathematical texts.
We set the model temperature to zero to make responses deterministic.
The maximum response length is 256 tokens.
Otherwise, we use standard generation parameters from the Hugging Face transformers library. We collected all responses in February 2024.

%%%%%%%%%%%%%%%%%%%%%%%%%%%%%%%%%%%%%%%%%%%%%%%%%%%%%%%%%
\subsection{Multinomial Logistic Regression}
\label{sec:reg}
We use simple multinomial logistic regression to classify the difficulty level of texts.
The task is to predict the difficulty level $C_{i}$ of a given educational text $S_{i}$.
$C_{i}$ can take three ordinal values: elementary, middle, or high school difficulty.
Instead of including $S_{i}$ directly, we include sets of \textsc{Static} and \textsc{Prompt-based} metrics $\textbf{M}_{i}$ that are computed based on $S_{i}$.
We regress $\textbf{M}_{i}$ on $C_{i}$ on the 3,638 training texts and then evaluate on the 910 test education texts.

We vary which metrics we include across experimental setups to evaluate the marginal benefits of different metrics.
There are three main setups of interest: 1) \textsc{Prompt-based} metrics only, 
2) \textsc{Static} metrics only, 
3) the combination of the two, which we refer to as \textsc{Combo}.

%%%%%%%%%%%%%%%%%%%%%%%%%%%%%%%%%%%%%%%%%%%%%%%%%%%%%%%%%
\subsection{Baseline: Zero- and Few-Shot Classification}

We exploit the general language capabilities of LLMs to compute \textsc{Prompt-based} metrics, which we then use as inputs to a logistic classifier for text difficulty.
A natural follow-up question is whether LLMs could directly predict text difficulty related to education levels. Therefore, we incorporate a baseline for zero-shot and few-shot text classification. We test zero-shot and few-shot classification with the same LLMs that we use for calculating our \textsc{Prompt-based} metrics. 
As an additional comparison point, we test GPT-4 Turbo.
%The promising capabilities of GPT-4, developed by OpenAI to tackle complex tasks, have been noted. However, due to the considerable expense associated with GPT-4, our study primarily focused on employing it for baseline classification tasks. Additionally, we test GPT-4 Turbo, featuring a context window of 128K. We access this model through its API.

 % Few-shot classifiers may outperform zero-shot ones, so we inquire into whether using a few-shot classifier with LLMs for predicting the education level of texts yields improvements over zero-shot experiments and compared to using a combination of \textsc{Prompt-based} metrics and \textsc{Static} metrics. 

Note that while the logistic classifier is fitted to our training data, the zero-shot LLM has not seen any examples at inference time. In the few-shot setting, we provide two examples for each education level and prompt the model to assign one of the desired labels without explanations.

To investigate the effect of different prompting styles, we test five distinct prompt templates in our zero-shot setup, each consisting of 25-30 words.
Additionally, each prompt contains a textual segment describing the text of the science question answering for educational-level classification.
We compare performance across the five prompt templates to determine the most effective prompt, i.e., the strongest baseline for our experiments. We evaluate the models' responses on a subset of randomly selected samples (n=100). The lowest performance stands at 29\%, while the highest achievement reaches 42\%. We proceed with our experiments under zero-shot and few-shot setups, using the best performance style as our baselines. The selected prompt for zero-shot experiments is:
``Your task is to predict the education level corresponding to a given text. You are provided with three labels to choose from: 
1) elementary school 
2) middle school 
3) high school.
Text: [text] Educational level: ''

% NOTE ON INVALID ANSWERS
We instructed LLMs to return one of the education levels. Due to the difficulty of LLMs in directly predicting the levels and complexity of the text, we have responses without the desired educational level. In this case, we assigned a default level to this invalid response, which is the ``elementary level''. For example, Llama2-13b has 2.86\% invalid in zero-shot and 4.07\% in few-shot. The most-predicted class is elementary school level, with 75.93\% in zero-shot and 80\% in few-shot. The number of invalid responses for other models is available in the Appendix~\ref{appendix-other-models}.

% We ask LLMs to classify the education level among labels: 1) elementary school, 2)middle school, and 3) highschool. Due to the difficulty of LLMs to directly predict the levels, we have responses without the desired educational level. In this case, we assigned a default level to this invalid response, which is the elementary level. Gemma7b has 10.33\% invalid response in zero-shot and 9.56\% over few-shot. The majority of the predicted class is high school level 73.41\% in zero shot and 72.75\% in few-shot. Mistral7b has 15.49\% invalid response in zero-shot and 6.37\% invalid in few-shot and with majority of classification for high school level with 66.04\% in zero-shot and 42.31\% in few-shot. Llama2-7b has 13.08\% invalid in zero-shot and 5.49\% in few-shot and majority of high school classification with 66.26\% in zero-shot and 76.04\% in few-shot. Llama2-13b has 2.86\% invalid in zero-shot and 4.07\% in few-shot. The majority of the predicted class is high school level with 75.93\% in zero-shot and 80\% in few-shot. Gpt-4 has only 5.93\% invalid in zero-shot and  0.77\% in few-shot. gpt-4 predicted also the high school level as the highest classification with 41.54\% in zero-shot and 40.22\% in few-shot.

%%%%%%%%%%%%%%%%%%%%%%%%%%%%%%%%%%%%%%%%%%%%%%%%%%%%%%%%%
\subsection{Results}
\label{sec:results}
\paragraph{Overall Performance}

Table \ref{tab:overall_performance_all_models} reports the overall results of our different logistic classifier setups along with the \textsc{Zero-shot} and \textsc{Few-shot} LLM classification baselines. We use Gemma-7b, Mistral-7b, Llama2-7b, and Llama2-13b across all referenced classification methods. GPT-4 is exclusively used in the baseline due to the high cost of experiments.

% New Table for all models
\begin{table*}
\centering
\begin{tabular}{llllll}
\toprule
\textbf{Method} & \textbf{Gemma-7b} & \textbf{Mistral-7b} & \textbf{Llama2-7b}  & \textbf{Llama2-13b} & \textbf{GPT-4}\\
\midrule
\textsc{Prompt-based} Reg. & 0.73 & 0.54 & 0.45 & 0.77 & - \\
\textsc{Static} Reg. & 0.81 & \textbf{0.81} & \textbf{0.81} & 0.81 & -\\
\textsc{Combo} Reg. & \textbf{0.95} & \textbf{0.82} & \textbf{0.81} & \textbf{0.88} & -\\
\midrule
\textsc{Zero-shot} LLM & 0.35 & 0.34 & 0.35 & 0.35 & 0.51\\
\textsc{Few-shot} LLM & 0.37 & 0.37 & 0.45 & 0.47 & \textbf{0.65} \\
\bottomrule
\end{tabular}
\caption{Macro-F1 for difficulty classification on test.
% Each column corresponds to one LLM.
\textsc{Prompt-based} metrics, zero-shot, and few-shot (two examples) performance are specific to each LLM.
\textsc{Static} metrics are the same across models.
Zero-shot and few-shot classification use GPT4. Best performance per model in \textbf{bold}.}
\label{tab:overall_performance_all_models}
\end{table*}

The findings highlight the consistent superiority of the \textsc{Combo} approach in achieving the highest macro-F1 score, surpassing all other models. Specifically, while the Llama2-7b model exhibits comparatively lower performance when employing the Prompt-based method, the Llama2-13b model demonstrates the best performance across \textsc{Prompt-based} metrics. Notably, the Gemma-7b model stands out as the best-performing model when using the \textsc{Combo} metric. In terms of Prompt-based regression, the average macro-F1 score across all models stands at 0.62, with all \textsc{Prompt-based} metrics obtained directly through LLMs' binary classification prompts. The best performance overall is achieved by \textsc{Combo}, which combines both sets of metrics, resulting in a macro-F1 score of 0.86.

% The results shows that the \textsc{Combo} approach consistently achieves the highest macro-F1 score, outperforming all other models. Specifically, while the Llama2-7b model demonstrates relatively lower performance when using the \textsc{Prompt-based} method, the Llama2-13b model exhibits better performance across Neural metrics. Notably, the Gemma-7b model stands out as the best-performing model when using the Combo metric. \textsc{Prompt-based} regression performs, on average 0.61 macro-F1 across all models. All \textsc{Prompt-based} metrics are collected directly by prompting the LLMs to make a binary classification.
% By comparison, \textsc{Static} regression performs substantially better, at average 0.81 macro-F1.
% Best overall is \textsc{Combo}, the combination of the two sets of metrics, at 0.86 macro-F1.

Nearly all models encounter difficulty in predicting the educational level across both \textsc{Zero-shot} and \textsc{Few-shot} methodologies. However, in these experiments, the \textsc{Few-shot} approach notably enhances the macro-F1 score. Additionally, Table \ref{tab:overall_performance_all_models} highlights that the best performance among baseline approaches is achieved by GPT-4, attaining a macro-F1 score of 0.63 in the \textsc{Few-shot} setting.

\paragraph{Performance by Education Level}
To delve into the performance more comprehensively, we split out the results for each regression setup by label, i.e., education level, in Table \ref{tab:label_performance}. Here, we display only the top-performing model based on the \textsc{Prompt-based} metric and provide the details of the other models in Appendix~\ref{appendix-other-models}.

\begin{table}[h]
    \centering
    \begin{tabularx}{0.47\textwidth}{llS[table-format=1.2] S[table-format=1.2] S[table-format=1.2]}
    \toprule
        & \textbf{Level} & \textbf{Precision} & \textbf{Recall} & \textbf{F1-Score}\\
        \midrule
        \multirow{3}{*}{\begin{sideways}\textsc{Prompt}\end{sideways}} & Elem. & 0.84 & 0.82 & 0.83 \\
        & Middle & 0.84 & 0.64 & 0.73 \\
        & High      & 0.68 & 0.84 & 0.75 \\
        \midrule
        \multirow{3}{*}{\begin{sideways}\textsc{Static}\end{sideways}} & Elem. & 0.86 & 0.85 & 0.86 \\
        & Middle     & 0.75 & 0.71 & 0.73\\
        & High      & \textbf{0.84} & 0.88 & 0.84 \\
        \midrule
        \multirow{3}{*}{\begin{sideways}\textsc{Combo}\end{sideways}} &  Elem. & \textbf{0.95}* & \textbf{0.93}* & \textbf{0.94}* \\
        & Middle & \textbf{0.89}* & \textbf{0.77}* & \textbf{0.83}* \\
        & High  & 0.82 &\textbf{0.93}* & \textbf{0.87}* \\
    \bottomrule

    \end{tabularx}
    \caption{Difficulty classification performance on test. $*=$ statistically significant improvements of \textsc{Combo} over \textsc{Static} at $p=0.05$ (bootstrap).
    \textsc{Prompt-based} metrics use \textit{Llama2-13b}.
    Best performance per level in \textbf{bold}.
    }
    \label{tab:label_performance}
\end{table}

The overall picture of \textsc{Prompt-based} regression shows that it faces difficulty in the classification of educational level, while \textsc{Static} performs much better, and \textsc{Combo} performs best, which indicates that there is an additional benefit to including the \textsc{Prompt-based} metrics.

We collect 1,000 bootstrap samples to train and test the logistic regression models for each approach. This method helps in understanding the variability and reliability of the model performance. We use t-tests to determine if the observed differences in accuracies are statistically significant over \textsc{Combo} vs. \textsc{Static}. Results in Table \ref{tab:label_performance} indicate a statistically significant improvement.

\paragraph{Feature Importance}
One big benefit of our regression approach over, for example, classification with an LLM, is that we can easily measure the feature importance of each metric that goes into the classification result.
For this purpose, we calculate univariate F-tests between each metric and the difficulty level variable.
Table \ref{tab:topfeatures} shows the top-five most important features, each among the \textsc{Prompt-based} and the \textsc{Static} metrics, based on these F-tests for \textit{Llama2-13b} model.

% Model:Llama-2-13B Table: Top selected features
\begin{table*}
\centering
\begin{tabularx}{0.99\textwidth}{llll}
\toprule
 & \textbf{Rank} & \textbf{Metric} & \textbf{F}  \\ 
\midrule
\multirow{5}{*}{\begin{sideways}Prompt Metrics\end{sideways}} & 1 & Based on the \textbf{ARI}, is this text suitable for ES readers?& 251.77* \\
& 2 & Is this text \textbf{relevant to curriculum} topics for ES students?& 249.07* \\
& 3 & Is this text about \textbf{math}? &  248.17* \\
& 4 & Is this text about \textbf{natural science}? & 240.07* \\
& 5 & Does this text contain \textbf{simple examples}? & 235.96* \\
\midrule
\multirow{5}{*}{\begin{sideways}Static Metrics\end{sideways}} & 1 & Gunning Fog (measures \textbf{readability}) & 817.86* \\
& 2 & Coleman-Liau index (measures \textbf{readability}) & 785.60* \\
& 3 & Flesch-Kincaid Reading Ease (measures \textbf{readability}) & 725.15* \\
& 4 & Automated Readability Index  (measures \textbf{Readability}) & 686.87*\\
& 5 & Number of unique Words (measures \textbf{lexical diversity}) & 613.89* \\
\bottomrule
\end{tabularx}
\caption{
Five most important features for \textsc{Prompt-based} and \textsc{Static} metrics in \textit{Llama2-13b}.
Feature importance is measured using univariate F-tests.
Larger F indicates higher feature importance. (ES: Elementary School, ARI: Automated Readability Index)
* indicates significance at >99.999\% confidence.
}
\label{tab:topfeatures}
\end{table*}

Most notably, the \textsc{Prompt-based} metrics are generally less important than the \textsc{Static} metrics. On average, the top five most important \textsc{Static} metrics are at least twice as significant as the top five \textsc{Prompt-based} metrics. The \textsc{Static} metrics mainly focus on readability and lexical diversity, while \textsc{Prompt-based} metrics capture topic relevancy and the inclusion of simple examples. Although they may not carry the same weight, all of the top metrics are highly statistically significant.

%%%%%%%%%%%%%%%%%%%%%%%%%%%%%%%%%%%%%%%%%%%%%%%%%%%%%%%%%%%%%%%%%%%%%%%%%%%%%%%%%%%%%%
%%%%%%%%%%%%%%%%%%%%%%%%%%%%%%%%%%%%%%%%%%%%%%%%%%%%%%%%%%%%%%%%%%%%%%%%%%%%%%%%%%%%%%
\section{Discussion}

%%%%%%%%%%%%%%%%%%%%%%%%%%%%%%%%%%%%%%%%%%%%%%%%%%%%%%%%%
\subsection{The Value of Prompt-based Metrics}

\textsc{Prompt-based} metrics by themselves may not be a good-enough basis for classifying text difficulty (Table~\ref{tab:overall_performance_all_models}).
\textsc{Static} metrics are much more effective by comparison.
However, our results also show that \textsc{Prompt-based} metrics do indeed capture relevant features of the text that are not captured by \textsc{Static} metrics since models that combine both kinds of metrics clearly perform best overall.
This is despite the fact that the \textsc{Static} metrics we include are many and highly diverse.

The practical usefulness of the particular \textsc{Prompt-based} metrics outlined in this paper is evident. Moreover, the broader application of \textsc{Prompt-based} metrics holds promise for evaluating text complexity. Our experiments indicate that the \textsc{Combo} approach outperforms other models consistently. Notably, most models exhibit superior macro-F1 scores in predicting elementary-level texts, suggesting that distinguishing science questions at the elementary level is more discernible compared to other educational levels.

Furthermore, we present the feature importance of \textsc{Prompt-based} metrics, noting that the primary \textsc{Prompt-based} metrics pertain to readability, understandability, and suitability of text for particular educational levels. Additionally, topic relevance (e.g., math or natural science) emerges as a significant feature. In top 5 best features of \textsc{Static} metrics are summarized through readability scores ranging from the Gunning Fog Index to the Flesch-Kincaid Index, along with a metric evaluating the lexical diversity of the text.

Better \textsc{Prompt-based} metrics identified in future work may be even more effective complements to Static metrics.

%%%%%%%%%%%%%%%%%%%%%%%%%%%%%%%%%%%%%%%%%%%%%%%%%%%%%%%%%
\subsection{Limitations}

\paragraph{Limited Scope of User Study}
The user study we conducted provides a clear empirical motivation for the \textsc{Prompt-based} metrics we selected.
This in itself is a core contribution of our work.
However, due to resource and time constraints, the sample of participants in the study is fairly small and of limited diversity.
Future work could improve on our approach by conducting larger studies or recruiting participants from even more relevant professions (e.g. teachers) to motivate the selection of even better \textsc{Prompt-based} metrics.

\paragraph{Limited Availability of Relevant Data}
Our experiments are mostly constrained by the availability of relevant data for text difficulty classification. 
The ScienceQA dataset that we use is, to our knowledge, the only dataset that fits our experimental setup in terms of size and detail on education level.
Therefore, we cannot make any strong claims about the generalisability of our results.
Future work could invest into building new datasets and testing cross-domain performance of both Static and \textsc{Prompt-based} metrics, which would give useful insights into which text features are most generally indicate of text difficulty.

%%%%%%%%%%%%%%%%%%%%%%%%%%%%%%%%%%%%%%%%%%%%%%%%%%%%%%%%%%%%%%%%%%%%%%%%%%%%%%%%%%%%%%
\section{Related Work}
\subsection{Question Answering Datasets in Education}

The review study by~\newcite{alkhuzaey2023text} about the literature on item difficulty classification reveals a significant shortage of publicly accessible datasets with items that are labeled according to their difficulty levels. For example,~\newcite{hsu2018automated} gathered their dataset from national standardized entrance tests that often concentrate on the medical and language fields, annotated with the performance data of 270,000 examinees. This study includes the necessity for a publicly accessible collection of standardized datasets and the need for further exploration into alternative methods for feature elicitation and classification modeling. The lack of publicly available datasets for measuring difficulty has led researchers toward the domain of Automatic Question Generation (AQG) in recent years. Typically, questions generated by AQG tend to be more straightforward in structure and cognitive demand than questions written by humans.

Most of these automatically generated questions are basic, primarily addressing only the first level of Bloom's taxonomy, which is focused on recall~\cite{leo2019ontology}. Another source of educational datasets is retrieved from online learning platforms or websites specific to the study's domain. An example includes the collection of 1,657 programming problems from LeetCode\footnote{\url{https://leetcode.com}}, labeled with the number of solutions submitted and the pass rate for each problem. Additionally, fewer datasets are from domain-specific textbooks and preparation books, particularly prevalent in the language domain for their role in training students for language proficiency exams. Domain experts developed the remaining sources to meet specific study goals, and according to~\newcite{alkhuzaey2023text}, only 7\% from school or university-level assessments.

The Stanford Question Answering Dataset (SQuAD), developed by~\newcite{rajpurkar2016squad}, features 150,000 questions in the form of paragraph-answer pairs sourced from Wikipedia articles. This dataset was utilized by~\newcite{bi2021simple} to develop and test their models for predicting the difficulty of reading comprehension questions.~\newcite{lu2022learn} created a multimodal science question-answering datasets, which includes 21,000 English passages from school reading exams, each accompanied by four multiple-choice questions. The ScienceQA dataset provides metadata fields for each question, including extensive solutions and general explanations which made it suitable for this study~\cite {lu2022learn}.

%%%%%%%%%%%%%%%%%%%%%%%%%%%%%%%%%%%%%%%%%%%%%%%%%%%%%%%%%
\subsection{Automatic Evaluation of Educational Content}

The difficulty level classification of questions presented to students is crucial for facilitating more effective and efficient learning. \newcite{perez2012automatic} shows teachers usually fail to identify the correct difficulty level of the questions according to their students' answers and final scores. The student's perception of the difficulty also changes across grades and subjects.~\newcite{alkhuzaey2023text} discovers that linguistic features significantly influence the determination of question difficulty levels in educational assessments. They have explored various syntactic and semantic aspects to understand the complexity of these questions. \newcite{crossley2019moving} shows the value of using crowdsourcing methods to gather human assessments of text comprehension, coupled with linguistic attributes derived from advanced readability metrics. This approach aids in creating models that explain how humans understand and process text, as well as factors influencing reading speed.~\newcite{crossley2023using} examined the effectiveness of new readability formulas developed on the CommonLit Ease of Readability (CLEAR) corpus using more efficient sentence-embedding models and comparing them to traditional readability formulas. They did not test LLMs directly for the difficulty classification task. In their respective studies, ~\newcite{imperial2023flesch}, ~\newcite{rooein2023know}, and ~\newcite{gobara2024llms} leverage Large Language Models (LLMs) for content generation, focusing specifically on controlling readability scores. Their research illuminates the inherent challenges and limitations encountered when attempting to effectively adapt LLMs for this purpose.

%%%%%%%%%%%%%%%%%%%%%%%%%%%%%%%%%%%%%%%%%%%%%%%%%%%%%%%%%

%%%%%%%%%%%%%%%%%%%%%%%%%%%%%%%%%%%%%%%%%%%%%%%%%%%%%%%%%%%%%%%%%%%%%%%%%%%%%%%%%%%%%%
%%%%%%%%%%%%%%%%%%%%%%%%%%%%%%%%%%%%%%%%%%%%%%%%%%%%%%%%%%%%%%%%%%%%%%%%%%%%%%%%%%%%%%
\section{Conclusion}
Good teachers succeed in making the material understandable for their respective audiences. This adaptation is a complex process that goes well beyond replacing individual words and phrases. However, existing \textsc{Static} metrics for text difficulty, like the Flesch-Kincaid Reading Ease score, still focus on precisely those elements. As a result, these metrics are crude and brittle, failing to adapt to new domains and working mainly on long-form documents.

Our experiments reveal the promising potential of LLMs in predicting educational difficulty through using the \textsc{Prompt-based} metrics rather than prompting the model directly. These metrics were derived from a small-scale user study involving students. Empirically, we demonstrate that when combined with traditional static metrics, these \textsc{Prompt-based} metrics enhance text difficulty classification. 

Our study paves the way for novel applications of LLMs in educational contexts. By involving more educational stakeholders, such as teachers, we can gather more representative \textsc{Prompt-based} metrics, facilitating future advancements in difficulty classification.

\section*{Ethical Considerations}
The participants in the user study we used in our paper were student volunteers for a course on related topics. They could leave the study at any point and were compensated in course credits that could be counted towards their study program. The study was conducted in accordance with the rules of the host university and passed its ethics assessment.
The risk for harm to the participants in this setting was assessed as minimal.

\section*{Acknowledgements}
Donya Rooein and Dirk Hovy were supported by the European Research Council (ERC) under the European Union’s Horizon 2020 research and innovation program (grant agreement No.\ 949944, INTEGRATOR). Paul Röttger was supported by a MUR FARE 2020 initiative under grant agreement Prot.\ R20YSMBZ8S (INDOMITA). They are members of the MilaNLP group and the Data and Marketing Insights Unit of the Bocconi Institute for Data Science and Analysis (BIDSA). Anastassia Shaitarova was supported by the National Centre of Competence in Research “Evolving Language”, Swiss National Science Foundation (SNSF) Agreement 51NF40 180888. 

% We would like to express our gratitude to the Department of Computational Linguistics at the University of Zurich for providing the facilities and time necessary for our practical experiments. Special thanks are due to the participating students, whose active engagement and insightful contributions greatly enhanced our research.

% \bibliographystyle{acl_natbib}
\bibliography{acl_latex.bib}

\begin{thebibliography}{30}
\expandafter\ifx\csname natexlab\endcsname\relax\def\natexlab#1{#1}\fi

\bibitem[{AlKhuzaey et~al.(2023)AlKhuzaey, Grasso, Payne, and Tamma}]{alkhuzaey2023text}
Samah AlKhuzaey, Floriana Grasso, Terry~R Payne, and Valentina Tamma. 2023.
\newblock Text-based question difficulty prediction: A systematic review of automatic approaches.
\newblock \emph{International Journal of Artificial Intelligence in Education}, pages 1--53.

\bibitem[{Baldwin et~al.(2021)Baldwin, Yaneva, Mee, Clauser, and Ha}]{baldwin2021using}
Peter Baldwin, Victoria Yaneva, Janet Mee, Brian~E Clauser, and Le~An Ha. 2021.
\newblock Using natural language processing to predict item response times and improve test construction.
\newblock \emph{Journal of Educational Measurement}, 58(1):4--30.

\bibitem[{Bi et~al.(2021)Bi, Cheng, Li, Qu, Shen, Qi, Pan, and Jiang}]{bi2021simple}
Sheng Bi, Xiya Cheng, Yuan-Fang Li, Lizhen Qu, Shirong Shen, Guilin Qi, Lu~Pan, and Yinlin Jiang. 2021.
\newblock Simple or complex? complexity-controllable question generation with soft templates and deep mixture of experts model.
\newblock \emph{arXiv preprint arXiv:2110.06560}.

\bibitem[{Chung et~al.(2022)Chung, Hou, Longpre, Zoph, Tay, Fedus, Li, Wang, Dehghani, Brahma et~al.}]{chung2022scaling}
Hyung~Won Chung, Le~Hou, Shayne Longpre, Barret Zoph, Yi~Tay, William Fedus, Yunxuan Li, Xuezhi Wang, Mostafa Dehghani, Siddhartha Brahma, et~al. 2022.
\newblock Scaling instruction-finetuned language models.
\newblock \emph{arXiv preprint arXiv:2210.11416}.

\bibitem[{Crossley et~al.(2023)Crossley, Choi, Scherber, and Lucka}]{crossley2023using}
Scott Crossley, Joon~Suh Choi, Yanisa Scherber, and Mathis Lucka. 2023.
\newblock Using large language models to develop readability formulas for educational settings.
\newblock In \emph{International Conference on Artificial Intelligence in Education}, pages 422--427. Springer.

\bibitem[{Crossley et~al.(2019)Crossley, Skalicky, and Dascalu}]{crossley2019moving}
Scott~A Crossley, Stephen Skalicky, and Mihai Dascalu. 2019.
\newblock Moving beyond classic readability formulas: New methods and new models.
\newblock \emph{Journal of Research in Reading}, 42(3-4):541--561.

\bibitem[{Flekova et~al.(2016)Flekova, Preo{\c{t}}iuc-Pietro, and Ungar}]{flekova2016exploring}
Lucie Flekova, Daniel Preo{\c{t}}iuc-Pietro, and Lyle Ungar. 2016.
\newblock Exploring stylistic variation with age and income on twitter.
\newblock In \emph{Proceedings of the 54th Annual Meeting of the Association for Computational Linguistics (Volume 2: Short Papers)}, pages 313--319.

\bibitem[{Flesch(1948)}]{flesch1948new}
Rudolph Flesch. 1948.
\newblock A new readability yardstick.
\newblock \emph{Journal of applied psychology}, 32(3):221.

\bibitem[{Gobara et~al.(2024)Gobara, Kamigaito, and Watanabe}]{gobara2024llms}
Seiji Gobara, Hidetaka Kamigaito, and Taro Watanabe. 2024.
\newblock Do llms implicitly determine the suitable text difficulty for users?
\newblock \emph{arXiv preprint arXiv:2402.14453}.

\bibitem[{Google(2024)}]{Google2024Gemma7b}
Google. 2024.
\newblock {Responsible Generative AI Toolk, GemmaTechnical Report}.
\newblock \url{https://ai.google.dev/gemma/docs}.
\newblock Accessed: March 6, 2024.

\bibitem[{Hosseini et~al.(2023)Hosseini, Gao, Liebovitz, Carvalho, Ahmad, Luo, MacDonald, Holmes, and Kho}]{hosseini2023exploratory}
Mohammad Hosseini, Catherine~A Gao, David~M Liebovitz, Alexandre~M Carvalho, Faraz~S Ahmad, Yuan Luo, Ngan MacDonald, Kristi~L Holmes, and Abel Kho. 2023.
\newblock An exploratory survey about using chatgpt in education, healthcare, and research.
\newblock \emph{medRxiv}, pages 2023--03.

\bibitem[{Hsu et~al.(2018)Hsu, Lee, Chang, and Sung}]{hsu2018automated}
Fu-Yuan Hsu, Hahn-Ming Lee, Tao-Hsing Chang, and Yao-Ting Sung. 2018.
\newblock Automated estimation of item difficulty for multiple-choice tests: An application of word embedding techniques.
\newblock \emph{Information Processing \& Management}, 54(6):969--984.

\bibitem[{Imperial and Madabushi(2023)}]{imperial2023flesch}
Joseph~Marvin Imperial and Harish~Tayyar Madabushi. 2023.
\newblock Flesch or fumble? evaluating readability standard alignment of instruction-tuned language models.
\newblock \emph{arXiv preprint arXiv:2309.05454}.

\bibitem[{Jiang et~al.(2023)Jiang, Sablayrolles, Mensch, Bamford, Chaplot, Casas, Bressand, Lengyel, Lample, Saulnier et~al.}]{jiang2023mistral}
Albert~Q Jiang, Alexandre Sablayrolles, Arthur Mensch, Chris Bamford, Devendra~Singh Chaplot, Diego de~las Casas, Florian Bressand, Gianna Lengyel, Guillaume Lample, Lucile Saulnier, et~al. 2023.
\newblock Mistral 7b.
\newblock \emph{arXiv preprint arXiv:2310.06825}.

\bibitem[{Leo et~al.(2019)Leo, Kurdi, Matentzoglu, Parsia, Sattler, Forge, Donato, and Dowling}]{leo2019ontology}
Jared Leo, Ghader Kurdi, Nicolas Matentzoglu, Bijan Parsia, Ulrike Sattler, Sophie Forge, Gina Donato, and Will Dowling. 2019.
\newblock Ontology-based generation of medical, multi-term mcqs.
\newblock \emph{International Journal of Artificial Intelligence in Education}, 29:145--188.

\bibitem[{Lin(2004)}]{lin-2004-rouge}
Chin-Yew Lin. 2004.
\newblock \href {https://aclanthology.org/W04-1013} {{ROUGE}: A package for automatic evaluation of summaries}.
\newblock In \emph{Text Summarization Branches Out}, pages 74--81, Barcelona, Spain. Association for Computational Linguistics.

\bibitem[{Lu et~al.(2022)Lu, Mishra, Xia, Qiu, Chang, Zhu, Tafjord, Clark, and Kalyan}]{lu2022learn}
Pan Lu, Swaroop Mishra, Tanglin Xia, Liang Qiu, Kai-Wei Chang, Song-Chun Zhu, Oyvind Tafjord, Peter Clark, and Ashwin Kalyan. 2022.
\newblock Learn to explain: Multimodal reasoning via thought chains for science question answering.
\newblock \emph{Advances in Neural Information Processing Systems}, 35:2507--2521.

\bibitem[{OpenAI(2023)}]{openai2023gpt4}
OpenAI. 2023.
\newblock \href {http://arxiv.org/abs/2303.08774} {{GPT-4 Technical Report}}.

\bibitem[{Papineni et~al.(2002)Papineni, Roukos, Ward, and Zhu}]{papineni-etal-2002-bleu}
Kishore Papineni, Salim Roukos, Todd Ward, and Wei-Jing Zhu. 2002.
\newblock \href {https://doi.org/10.3115/1073083.1073135} {{B}leu: a method for automatic evaluation of machine translation}.
\newblock In \emph{Proceedings of the 40th Annual Meeting of the Association for Computational Linguistics}, pages 311--318, Philadelphia, Pennsylvania, USA. Association for Computational Linguistics.

\bibitem[{P{\'e}rez et~al.(2012)P{\'e}rez, Santos, P{\'e}rez, de~Castro~Fern{\'a}ndez, and Mart{\'\i}n}]{perez2012automatic}
Elena~Verd{\'u} P{\'e}rez, Luisa M~Regueras Santos, Mar{\'\i}a Jes{\'u}s~Verd{\'u} P{\'e}rez, Juan~Pablo de~Castro~Fern{\'a}ndez, and Ricardo~Garc{\'\i}a Mart{\'\i}n. 2012.
\newblock Automatic classification of question difficulty level: Teachers' estimation vs. students' perception.
\newblock In \emph{2012 Frontiers in Education Conference Proceedings}, pages 1--5. IEEE.

\bibitem[{Rajpurkar et~al.(2016)Rajpurkar, Zhang, Lopyrev, and Liang}]{rajpurkar2016squad}
Pranav Rajpurkar, Jian Zhang, Konstantin Lopyrev, and Percy Liang. 2016.
\newblock Squad: 100,000+ questions for machine comprehension of text.
\newblock \emph{arXiv preprint arXiv:1606.05250}.

\bibitem[{Recasens and Hovy(2011)}]{recasens2011blanc}
Marta Recasens and Eduard Hovy. 2011.
\newblock Blanc: Implementing the rand index for coreference evaluation.
\newblock \emph{Natural language engineering}, 17(4):485--510.

\bibitem[{Rooein et~al.(2023)Rooein, Curry, and Hovy}]{rooein2023know}
Donya Rooein, Amanda~Cercas Curry, and Dirk Hovy. 2023.
\newblock Know your audience: Do llms adapt to different age and education levels?
\newblock \emph{arXiv preprint arXiv:2312.02065}.

\bibitem[{Sallam(2023)}]{healthcare11060887}
Malik Sallam. 2023.
\newblock \href {https://doi.org/10.3390/healthcare11060887} {Chatgpt utility in healthcare education, research, and practice: Systematic review on the promising perspectives and valid concerns}.
\newblock \emph{Healthcare}, 11(6).

\bibitem[{Stamps(2004)}]{stamps2004effectiveness}
Lisa~S Stamps. 2004.
\newblock The effectiveness of curriculum compacting in first grade classrooms.
\newblock \emph{Roeper Review}, 27(1):31--41.

\bibitem[{Touvron et~al.(2023)Touvron, Martin, Stone, Albert, Almahairi, Babaei, Bashlykov, Batra, Bhargava, Bhosale et~al.}]{touvron2023llama}
Hugo Touvron, Louis Martin, Kevin Stone, Peter Albert, Amjad Almahairi, Yasmine Babaei, Nikolay Bashlykov, Soumya Batra, Prajjwal Bhargava, Shruti Bhosale, et~al. 2023.
\newblock Llama 2: Open foundation and fine-tuned chat models.
\newblock \emph{arXiv preprint arXiv:2307.09288}.

\bibitem[{Upadhyay et~al.(2023)Upadhyay, Ginsberg, and Callison-Burch}]{upadhyay2023improving}
Shriyash Upadhyay, Etan Ginsberg, and Chris Callison-Burch. 2023.
\newblock Improving mathematics tutoring with a code scratchpad.
\newblock In \emph{Proceedings of the 18th Workshop on Innovative Use of NLP for Building Educational Applications (BEA 2023)}, pages 20--28.

\bibitem[{Xue et~al.(2020)Xue, Yaneva, Runyon, and Baldwin}]{xue2020predicting}
Kang Xue, Victoria Yaneva, Christopher Runyon, and Peter Baldwin. 2020.
\newblock Predicting the difficulty and response time of multiple choice questions using transfer learning.
\newblock In \emph{Proceedings of the Fifteenth Workshop on Innovative Use of NLP for Building Educational Applications}, pages 193--197.

\bibitem[{Yan et~al.(2023)Yan, Sha, Zhao, Li, Martinez-Maldonado, Chen, Li, Jin, and Ga{\v{s}}evi{\'c}}]{yan2023practical}
Lixiang Yan, Lele Sha, Linxuan Zhao, Yuheng Li, Roberto Martinez-Maldonado, Guanliang Chen, Xinyu Li, Yueqiao Jin, and Dragan Ga{\v{s}}evi{\'c}. 2023.
\newblock Practical and ethical challenges of large language models in education: A systematic literature review.
\newblock \emph{arXiv preprint arXiv:2303.13379}.

\bibitem[{Yaneva et~al.(2019)Yaneva, Baldwin, Mee et~al.}]{yaneva2019predicting}
Victoria Yaneva, Peter Baldwin, Janet Mee, et~al. 2019.
\newblock Predicting the difficulty of multiple choice questions in a high-stakes medical exam.
\newblock In \emph{Proceedings of the Fourteenth Workshop on Innovative Use of NLP for Building Educational Applications}, pages 11--20.

\end{thebibliography}

\appendix

\section{Selected Prompts from the User Study}
\label{appendix-users}
We collect the top prompts of the students from the chat history with analytical, manual, and AI Assistant (ChatGPT).
\subsection{Elementary School:}
- Simplify a text for elementary school, using simple language for 6-12 years olds.
- Create an elementary version of a high school lecture text.
- Simplify a high school text for elementary school.
- Explain in a way an 8-year-old would understand.
- This is a text meant for high school students. Can you help me make an appropriate version for elementary school students with very simple language and comprehensive, easy-to-understand examples?

\subsection{Middle School:}

- Give examples from middle school lectures.
- Adapt a high school text for middle school, using less advanced language.
- Be more textbook-like and more to the point for the level of middle school.
- Adapt content for a student in middle school.
- Simplify a lecture text for middle school using illustrative examples.

\subsection{High School:}

- Enhance scientific accuracy and add comprehensive examples for the high school level.
- Adapt a middle school text for high school, using advanced language.
- Increase difficulty for high school, with advanced vocabulary and scientific concepts.
- Can you make it more scientific and less story-telling-like?
- Increase the difficulty level with comprehensive examples.

\section{Parameter settings}
The Static metrics are collected by Python packages such as \texttt{nltk(3.8.1)}, \texttt{pandas(2.2.0)}, \texttt{textstat(0.7.3}, \texttt{spacy(3.7.4)}. We use \texttt{nltk.download} to get data for `stopwords', `cmudict', `wordnet', and `averaged\_perceptron\_tagger'. 

We performed 8-bit quantization for collecting prompt-based metrics and baselines, with a maximum input length of 2048 tokens and a maximum output length of 256 tokens. This process was restricted to a single run due to our utilization of pre-trained models readily accessible in HuggingFace Transformers.

For Regression model, we use scikit-learn package and SelectKBest with \texttt{f\_classif} score function.

\section{List of Metrics}
\label{appendix-prompts}

%%%%%%%%%%%%%%%%%%%%%%%%%%%%
\subsection{Static Metrics}
Table~\ref{tab:features} shows all static metrics.
\begin{table*}[htbp]
    \centering
    \caption{List of Static metrics}
    \begin{tabular}{|l|l|}
        \hline
        \textbf{Feature} & \textbf{Description} \\
        \hline
        \texttt{n\_words\_q} & Number of words in the question \\
        \texttt{n\_words\_a\_solution} & Number of words in the solution of an answer \\
        \texttt{n\_words\_a\_lecture} & Number of words in the lecture \\
        \texttt{Text\_Length} & Length of the text \\
        \texttt{Word\_Count} & Total word count \\
        \texttt{Nouns} & Number of nouns \\
        \texttt{Verbs} & Number of verbs \\
        \texttt{Adjectives} & Number of adjectives \\
        \texttt{Adverbs} & Number of adverbs \\
        \texttt{Num\_Numbers} & Number of numeric characters \\
        \texttt{Num\_Commas} & Number of commas \\
        \texttt{Num\_Complex\_Words} & Number of complex words \\
        \texttt{Num\_Unique\_Words} & Number of unique words \\
        \texttt{Num\_Content\_Words} & Number of content words \\
        \texttt{Num\_Content\_Words\_No\_Stopwords} & Number of content words excluding stopwords \\
        \texttt{Word\_Length\_Syllables} & Average word length in syllables \\
        \texttt{Avg\_Sentence\_Length} & Average sentence length \\
        \texttt{Num\_Prepositional\_Phrases} & Number of prepositional phrases \\
        \texttt{Num\_Negated\_Words\_Stem} & Number of negated words stemmed \\
        \texttt{Num\_Negated\_Words\_Lead\_In} & Number of negated words leading in \\
        \texttt{Num\_Main\_Noun\_Phrases} & Number of main noun phrases \\
        \texttt{Avg\_Main\_NP\_Length} & Average length of main noun phrases \\
        \texttt{Num\_Verb\_Phrases} & Number of verb phrases \\
        \texttt{Prop\_Active\_Voice\_Verbs} & Proportion of active voice verbs \\
        \texttt{Prop\_Passive\_Voice\_Verbs} & Proportion of passive voice verbs \\
        \texttt{Ratio\_Active\_to\_Passive\_Verbs} & Ratio of active to passive voice verbs \\
        \texttt{Num\_Words\_Before\_Main\_Verb} & Number of words before the main verb \\
        \texttt{Num\_Agentless\_Passive\_Constructions} & Number of agentless passive constructions \\
        \texttt{Word\_Length\_Std\_Dev} & Standard deviation of word lengths \\
        \texttt{Num\_Polysemic\_Words} & Number of polysemic words \\
        \texttt{Num\_Word\_Senses} & Number of word senses \\
        \texttt{Num\_Word\_Senses\_For\_Content\_Words} & Number of word senses for content words \\
        \texttt{Num\_Word\_Senses\_For\_Nouns} & Number of word senses for nouns \\
        \texttt{Num\_Word\_Senses\_For\_Verbs} & Number of word senses for verbs \\
        \texttt{Num\_Word\_Senses\_For\_Non\_Auxiliary\_Verbs} & Number of word senses for non-auxiliary verbs \\
        \texttt{Num\_Word\_Senses\_For\_Adjectives} & Number of word senses for adjectives \\
        \texttt{Num\_Word\_Senses\_For\_Adverbs} & Number of word senses for adverbs \\
        \texttt{Distance\_To\_Root\_Nouns} & Distance to root for nouns \\
        \texttt{Distance\_To\_Root\_Verbs} & Distance to root for verbs \\
        \texttt{flesch\_kincaid\_grade} & Flesch-Kincaid grade level \\
        \texttt{flesch\_kincaid\_ease} & Flesch-Kincaid ease score \\
        \texttt{coleman\_liau\_index} & Coleman-Liau index \\
        \texttt{automated\_readability\_index} & Automated Readability Index \\
        \texttt{smog\_index} & SMOG index \\
        \texttt{gunning\_fog} & Gunning Fog index \\
        \texttt{traenkle\_bailer\_index} & Traenkle-Bailer index \\
        \hline
    \end{tabular}
    \label{tab:features}
\end{table*}

\subsection{Prompt-based Metrics}
Is this text readable for an elementary school student?,  Is this text suitable for an elementary school student?,  Is this text easy to understand for elementary school students?,  Is this text relevant to curriculum topics for elementary school students?,  Is this text relevant to the knowledge and experiences of elementary school students?,  Could an average elementary school student engage with the content of this task?,  Could most elementary school students complete this task without significant difficulty?,  Is this text appropriate for the skills and knowledges of elementary school students?,  Is the length of this text suitable for elementary school students?,  Would the vocabulary in this text be comprehensible to elementary school students?,  Is this text readable for a middle school student?,  Is this text suitable for a middle school student?,  Is this text easy to understand for middle school students?,  Is this text relevant to curriculum topics for middle school students?,  Is this text relevant to the knowledge and experiences of middle school students?,  Could an average middle school student engage with the content of this task?,  Could most middle school students complete this task without significant difficulty?,  Is this text appropriate for the skills and knowledges of middle school students?,  Is the length of this text suitable for middle school students?,  Would the vocabulary in this text be comprehensible to middle school students?,  Is this text readable for a high school student?,  Is this text suitable for a high school student?,  Is this text easy to understand for high school students?,  Is this text relevant to curriculum topics for high school students?,  Is this text relevant to the knowledge and experiences of high school students?,  Could an average high school school student engage with the content of this task?,  Could most high school students complete this task without significant difficulty?,  Is this text appropriate for the skills and knowledges of high school students?,  Is the length of this text suitable for high school students?,  Would the vocabulary in this text be comprehensible to high school students?,  Does this text contain metaphors and/or figurative language?,  Does this text use complex language?,  Does this text use simple language?,  Does this text contain technical jargon?,  Is this text about science?,  Is this text about language science?,  Is this text about natural science?,  Is this text about social science?,  Is this text about math?,  Is this text about physics?,  Is this text about chemistry?,  Is this text about earth science?,  Is this text about world history?,  Is this text about geography?,  Based on the Flesch-Kincaid reading-ease score, is this text suitable for elementary school readers?,  Based on the Flesch-Kincaid reading-ease score, is this text suitable for middle school readers?,  Based on the Flesch-Kincaid reading-ease score, is this text suitable for high school readers?,  Based on the Gunning Fog Index, is this text suitable for elementary school readers?,  Based on the Gunning Fog Index, is this text suitable for middle school readers?,  Based on the Gunning Fog Index, is this text suitable for high school readers?,  Based on the Coleman-Liau Index, is this text suitable for elementary school readers?,  Based on the Coleman-Liau Index, is this text suitable for middle school readers?,  Based on the Coleman-Liau Index, is this text suitable for high school readers?,  Based on the Automated Readability Index (ARI), is this text suitable for elementary school readers?,  Based on the Automated Readability Index (ARI), is this text suitable for middle school readers?,  Based on the Automated Readability Index (ARI), is this text suitable for high school readers?,  Based on the SMOG Index, is this text suitable for elementary school readers?,  Based on the SMOG Index, is this text suitable for middle school readers?,  Based on the SMOG Index, is this text suitable for high school readers?,  Does this text contain basic concepts that are easy to comprehend?,  Does this text cover multiple concepts?,  Does this text provide a very explicit explanation?,  Does this text contain simple examples?

\section{Details over Gemma-7B, Mistral-7B, and Llama2-7B }
\label{appendix-other-models}
We describe the performance of these models in detail. Gemma7b has 10.33\% invalid response in zero-shot and 9.56\% over few-shot. The majority of the predicted class is high school level 73.41\% in zero-shot and 72.75\% in few-shot. Mistral7b has 15.49\% invalid response in zero-shot and 6.37\% invalid in few-shot and with majority of classification for high school level with 66.04\% in zero-shot and 42.31\% for elemetary school in few-shot. Llama2-7b has 13.08\% invalid in zero-shot and 5.49\% in few-shot and the majority of elementary school classification with 66.26\% in zero-shot and also 76.04\% in few-shot. Gpt-4 has only 5.93\% invalid in zero-shot and  0.77\% in few-shot. Gpt-4 predicted also the high school level as the highest classification with 41.54\% in zero-shot and 40.22\% in few-shot.

% Table: Overal Accuracy Results
%%%%%%%%%%%%%%%%%%%%%%%%%%%%%%%%%%%%%%%%%%
% Model "Gemma-7b-it"
%%%%%%%%%%
\begin{table}[h]
    \centering
    \begin{tabularx}{0.47\textwidth}{llS[table-format=1.2] S[table-format=1.2] S[table-format=1.2]}
    \toprule
        & \textbf{Level} & \textbf{Precision} & \textbf{Recall} & \textbf{F1-Score}\\
        \midrule
        \multirow{3}{*}{\begin{sideways}\textsc{Prompt}\end{sideways}} & Elem. & 0.83 & 0.81 & 0.82 \\
        & Middle & 0.75 & 0.57 & 0.65 \\
        & High      & 0.66 & 0.81 & 0.65 \\
        \midrule
        \multirow{3}{*}{\begin{sideways}\textsc{Static}\end{sideways}} & Elem. & 0.86 & 0.85 & 0.86 \\
        & Middle     & 0.75 & 0.71 & 0.73\\
        & High      & 0.84 & 0.88 & 0.86 \\
        \midrule
        \multirow{3}{*}{\begin{sideways}\textsc{Combo}\end{sideways}} &  Elem. & \textbf{0.98}* & \textbf{0.98}* & \textbf{0.98}* \\
        & Middle & \textbf{0.98}* & \textbf{0.91}* & \textbf{0.95}* \\
        & High  & \textbf{0.91}* &\textbf{0.97}* & \textbf{0.94}* \\
    \bottomrule
    \end{tabularx}
  \caption{Difficulty classification performance on test. $*=$ statistically significant improvements of \textsc{Combo} over \textsc{Static} at $p=0.05$ (bootstrap).
    \textsc{Prompt-based} metrics use \textit{Gemma-7b}.
    Best performance per level in \textbf{bold}.
    }
    \label{tab:label_performance-Gemma7b}
\end{table}

%%%%%%%%%%%%%%%%%%%%%%%%%%%%%%%%%%%%%%%%%%%
% Model" Mistral-7B, 

\begin{table}[h]
    \centering
    \begin{tabularx}{0.47\textwidth}{llS[table-format=1.2] S[table-format=1.2] S[table-format=1.2]}
    \toprule
        & \textbf{Level} & \textbf{Precision} & \textbf{Recall} & \textbf{F1-Score}\\
        \midrule
        \multirow{3}{*}{\begin{sideways}\textsc{Prompt}\end{sideways}} & Elem. & 0.46 & 0.86 & 0.60 \\
        & Middle & \textbf{0.92} & 0.83 & \textbf{0.88} \\
        & High      & 0.34 & 0.10  & 0.16 \\
        \midrule
        \multirow{3}{*}{\begin{sideways}\textsc{Static}\end{sideways}} & Elem. & \textbf{0.86} & 0.85 & 0.86 \\
        & Middle     & 0.75 & 0.71 & 0.73\\
        & High      & 0.84 & \textbf{0.88} & \textbf{0.86} \\
        \midrule
        \multirow{3}{*}{\begin{sideways}\textsc{Combo}\end{sideways}} &  Elem. & 0.76 & \textbf{0.95}* & \textbf{0.84}* \\
        & Middle & 0.85 & \textbf{0.90}* & \textbf{0.88}* \\
        & High  & \textbf{0.89 }* & 0.64 & 0.75 \\
    \bottomrule
    \end{tabularx}
 \caption{Difficulty classification performance on test. $*=$ statistically significant improvements of \textsc{Combo} over \textsc{Static} at $p=0.05$ (bootstrap).
    \textsc{Prompt-based} metrics use \textit{Mistral-7b}.
    Best performance per level in \textbf{bold}.
    }
    \label{tab:label_performance-mistral}
\end{table}
%%%%%%%%%%%%%%%%%%%%%%%%%%%%%%%%%%%%%%%%%%%%%%%%%
% Model" llama-7B, 

\begin{table}[h]
    \centering
    \begin{tabularx}{0.47\textwidth}{llS[table-format=1.2] S[table-format=1.2] S[table-format=1.2]}
    \toprule
        & \textbf{Level} & \textbf{Precision} & \textbf{Recall} & \textbf{F1-Score}\\
        \midrule
        \multirow{3}{*}{\begin{sideways}\textsc{Prompt}\end{sideways}} & Elem. & 0.44 & 0.47 & 0.45 \\
        & Middle & 0.62 & 0.61 & 0.62 \\
        & High   & 0.29 & 0.28 & 0.28 \\
        \midrule
        \multirow{3}{*}{\begin{sideways}\textsc{Static}\end{sideways}} & Elem. & 0.86 & 0.85 & 0.86 \\
        & Middle     & \textbf{0.75} & 0.71 & \textbf{0.73}\\
        & High      & \textbf{0.84} & \textbf{0.88} & \textbf{0.86} \\
        \midrule
        \multirow{3}{*}{\begin{sideways}\textsc{Combo}\end{sideways}} &  Elem. & \textbf{0.88}* & \textbf{0.97}* & \textbf{0.93}* \\
        & Middle & 0.72 & \textbf{0.74}* & \textbf{0.73}* \\
        & High  & 0.83& 0.73 & 0.78 \\
    \bottomrule
    \end{tabularx}
\caption{Difficulty classification performance on test. $*=$ statistically significant improvements of \textsc{Combo} over \textsc{Static} at $p=0.05$ (bootstrap).
    \textsc{Prompt-based} metrics use \textit{Llama2-7b}.
    Best performance per level in \textbf{bold}.
    }
    \label{tab:label_performance-llama7b}
\end{table}

% %%%%%%%%%%%%%%%%%%%%%%%%%%%%%%%%%%%%%%%%%%%%%%%%%%%%%%%%

% Model:Gemma-7B Table: Top selected features
\begin{table*}
\centering
\begin{tabularx}{0.99\textwidth}{llll}
\toprule
 & \textbf{Rank} & \textbf{Metric} & \textbf{F}  \\ 
\midrule
\multirow{5}{*}{\begin{sideways}Prompt\end{sideways}} & 1 &Based on the \textbf{Coleman-Liau Index}, is the text suitable for MS readers?& 105.09* \\
& 2 &  Is this text \textbf{readable} for a MS student? & 104.42* \\
& 3 & Based on the \textbf{SMOG Index}, is this text suitable for MS readers? &  103.53* \\
& 4 & Is this text \textbf{suitable} for a MS student? & 94.21* \\
& 5 & Based on the \textbf{Gunning Fog Index}, is this text suitable for MS readers? & 92.35* \\
\midrule
\multirow{5}{*}{\begin{sideways}Static Metrics\end{sideways}} & 1 & Gunning Fog (measures text \textbf{readability}) & 817.86* \\
& 2 & Coleman-Liau index (measures text \textbf{readability}) & 785.60* \\
& 3 & Flesch-Kincaid Reading Ease (measures \textbf{readability}) & 725.15* \\
& 4 & Automated Readability Index  (measures \textbf{lexical diversity}) & 686.87*\\
& 5 & Number of unique Words (measures \textbf{lexical diversity}) & 613.89* \\
\bottomrule
\end{tabularx}
\caption{
Top five most important features among the \textsc{Prompt-based} and \textsc{Static} metrics.
Feature importance is measured using univariate F-tests.
Larger F indicates higher feature importance. (MS: Middle School)
\textsc{Prompt-based} metrics use the \textit{Gemma-7B} model.
* indicates significance at >99.999\% confidence. 
}
\label{tab:topfeatures-gemma7b}
\end{table*}

% %%%%%%%%%%%%%%%%%%%%%%%%%%%%%%%%%%%%%%%%%%%%%%%%%%%%%%%%

% Model:Mistral-7B Table: Top selected features
\begin{table*}
\centering
\begin{tabularx}{0.99\textwidth}{llll}
\toprule
 & \textbf{Rank} & \textbf{Metric} & \textbf{F}  \\ 
\midrule
\multirow{5}{*}{\begin{sideways}Prompt\end{sideways}} & 1 &Based on the \textbf{Gunning Fog Index}, is this text suitable for ES readers?& 209.84* \\
& 2 & Is this text \textbf{easy to understand} for ES students??  & 193.22* \\
& 3 & Is this text \textbf{Suitable} for ES students & 190.61* \\
& 4 & Is this text about \textbf{math}? & 175.72* \\
& 5 & Is this text \textbf{relevant to curriculum} topics for ES students? & 175.08* \\
\midrule
\multirow{5}{*}{\begin{sideways}Static Metrics\end{sideways}} & 1 & Gunning Fog (measures text \textbf{readability}) & 817.86* \\
& 2 & Coleman-Liau index (measures text \textbf{readability}) & 785.60* \\
& 3 & Flesch-Kincaid Reading Ease (measures \textbf{readability}) & 725.15* \\
& 4 & Automated Readability Index  (measures \textbf{Readability}) & 686.87*\\
& 5 & Number of unique Words (measures \textbf{lexical diversity}) & 613.89* \\
\bottomrule
\end{tabularx}
\caption{
Top five most important features among the \textsc{Prompt-based} and \textsc{Static} metrics.
Feature importance is measured using univariate F-tests.
Larger F indicates higher feature importance. (ES: Elementary School)
\textsc{Prompt-based} metrics use the \textit{Mistral-7B} model.
* indicates significance at >99.999\% confidence. 
}
\label{tab:topfeatures-mistal7b}
\end{table*}

% -------------------------------------------------------------

% Model:Llama7b Table: Top selected features
\begin{table*}
\centering
\begin{tabularx}{0.99\textwidth}{llll}
\toprule
 & \textbf{Rank} & \textbf{Metric} & \textbf{F}  \\ 
\midrule
\multirow{5}{*}{\begin{sideways}Prompt-based Metrics\end{sideways}} & 1 & Is this text \textbf{relevant to curriculum} topics for ES students?& 139.66* \\
& 2 & Is this text suitable for an ES student? & 136.97* \\
& 3 & Is this text \textbf{readable} for an ES student  & 132.89* \\
& 4 & Based on the \textbf{Gunning Fog Index}, is this text \textbf{suitable} for MS readers?" & 125.51* \\
& 5 &  Is this text about \textbf{natural science}? & 124.52* \\
\midrule
\multirow{5}{*}{\begin{sideways}Static Metrics\end{sideways}} & 1 & Gunning Fog (measures text \textbf{readability}) & 817.86* \\
& 2 & Coleman-Liau index (measures text \textbf{readability}) & 785.60* \\
& 3 & Flesch-Kincaid Reading Ease (measures \textbf{readability}) & 725.15* \\
& 4 & Automated Readability Index  (measures \textbf{Readability}) & 686.87*\\
& 5 & Number of unique Words (measures \textbf{lexical diversity}) & 613.89* \\
\bottomrule
\end{tabularx}
\caption{
Top five most important features among the \textsc{Prompt-based} and \textsc{Static} metrics.
Feature importance is measured using univariate F-tests.
Larger F indicates higher feature importance.(ES: Elementary School, MS: Middle School)
\textsc{Prompt-based} metrics use the \textit{Llamma2-7B} model.
* indicates significance at >99.999\% confidence. 
}
\label{tab:topfeatures-llama7b}
\end{table*}

%%%%%%%%%%%%%%%%%%%%%%%%%%%%%%%%%%%%%%%%%%%%%%%%%%%%%%%

\end{document}